\documentclass[a4paper,twoside]{article}

\usepackage{epsfig}
\usepackage{subfigure}
\usepackage{calc}
\usepackage{amssymb}
\usepackage{amstext}
\usepackage{amsmath}
\usepackage{amsthm}
\usepackage{multicol}
\usepackage{pslatex}
\usepackage{apalike}

\usepackage{SCITEPRESS}     % Please add other packages that you may need BEFORE the SCITEPRESS.sty package.

\begin{document}

\title{Automatic Viseme Vocabulary Construction to Enhance \\ Continuous Lip-reading}
%Automatic viseme vocabulary construction to enhance continuous lip-reading
%Automatic Phoneme to Viseme Mapping to Enhance \hspace{50mm} Visual-only Speech Recognition
\author{\authorname{Adriana Fernandez-Lopez\sup{1} and Federico M. Sukno\sup{1}}
\affiliation{\sup{1}Department of Information and Communication Technologies, Pompeu Fabra University, Barcelona, Spain}
\email{\{adriana.fernandez, federico.sukno\}@upf.edu}
}


\abstract{Speech is the most common communication method between humans and involves the perception of both auditory and visual channels. Automatic speech recognition focuses on interpreting the audio signals, but it has been demonstrated that video can provide information that is complementary to the audio. Thus, the study of automatic lip-reading is important and is still an open problem. One of the key challenges is the definition of the visual elementary units (the visemes) and their vocabulary. Many researchers have analyzed the importance of the phoneme to viseme mapping and have proposed viseme vocabularies with lengths between 11 and 15 visemes. These viseme vocabularies have usually been manually defined by their linguistic properties and in some cases using decision trees or clustering techniques. In this work, we focus on the automatic construction of an optimal viseme vocabulary based on the association of phonemes with similar appearance. To this end, we construct an automatic system that uses local appearance descriptors to extract the main characteristics of the mouth region and HMMs to model the statistic relations of both viseme and phoneme sequences. To compare the performance of the system different descriptors (PCA, DCT and SIFT) are analyzed. We test our system in a Spanish corpus of continuous speech. Our results indicate that we are able to recognize approximately 58\% of the visemes, 47\% of the phonemes and 23\% of the words in a continuous speech scenario and that the optimal viseme vocabulary for Spanish is composed by 20 visemes.}

\onecolumn \maketitle \normalsize \vfill

\section{\uppercase{Introduction}}
\label{sec:intro}
Speech is the most used communication method between humans, and it is considered a multi-sensory process that involves perception of both acoustic and visual cues since McGurk demonstrated the influence of vision in speech perception \cite{mcgurk1976hearing}. Many authors have subsequently demonstrated that the incorporation of visual information into speech recognition systems improves robustness \cite{potamianos2003recent}.
Much of the research in automatic speech recognition (ASR) systems has focused on audio speech recognition, or on the combination of both modalities using audiovisual speech recognition (AV-ASR) systems to improve the recognition rates, but visual automatic speech recognition systems (VASR) are rarely analyzed alone \cite{dupont2000audio}, \cite{nefian2002coupled}, \cite{zhou2014review}, \cite{yau2007visual}.

Even though the audio is in general much more informative than the video signal, human speech perception relies on the visual information to help decoding spoken words as auditory conditions are degraded \cite{erber1975auditory}, \cite{sumby1954visual}, \cite{hilder2009comparison}, \cite{ronquest2010language}. In addition visual information provides complementary information as speaker localization, articulation place, and the visibility of the tongue, the teeth and the lips. Furthermore, for people with hearing impairments, the visual channel is the only source of information if there is no sign language interpreter \cite{seymour2008comparison}, \cite{potamianos2003recent}, \cite{antonakos2015survey}.

The performance of audio only ASR systems is very high if there is not much noise to degrade the signal. However, in noisy environments AV-ASR systems improve the recognition performance when compared to their audio-only equivalents \cite{potamianos2003recent}, \cite{dupont2000audio}. On the contrary, in visual only ASR systems the recognition rates are rather low. It is true that the access to speech recognition through the visual channel is subject to a series of limitations. One of the key limitations relies on the ambiguities that arise when trying to map visual information into the basic phonetic unit (the phonemes), i.e. not all the phonemes that are heard can be distinguished by observing the lips. There are two types of ambiguities: $i)$ there are phonemes that are easily confused because they are perceived visually similar to others. For example, the phones /p/ and /b/ are visually indistinguishable because voicing occurs at the glottis, which is not visible. $ii)$ there are phonemes whose visual appearance can change (or even disappear) depending on the context (co-articulated consonants). This is the case of the \emph{velars}, consonants articulated with the back part of the tongue against the soft palate (e.g: /k/ or /g/), because they change their position in the palate depending on the previous or following phoneme  \cite{moll1971investigation}. In consequence of these limitations, there is no one-to-one mapping between the phonetic transcription of an utterance and their corresponding visual transcription \cite{chictu12012automatic}. From the technical point of view lip-reading depends on the distance between the speakers, on the illumination conditions and on the visibility of the mouth \cite{hilder2009comparison}, \cite{buchan2007spatial}, \cite{ortiz2008lipreading}.

The objective of ASR systems is to recognize words. Words can be represented as strings of phonemes, which can then be mapped to acoustic observations using pronunciation dictionaries that establish the mapping between words and phonemes. In analogy to audio speech systems, where there is consensus that the phoneme is the standard minimal unit for speech recognition, when adding visual information we aim at defining \emph{visemes}, namely the minimum distinguishable speech unit in the video domain \cite{fisher1968confusions}. As explained above, the mapping from phonemes to visemes cannot be one-to-one, but apart from this fact there is no much consensus on their definition nor in their number. When designing VASR systems, one of the most important challenges is the viseme vocabulary definition. There are discrepancies on whether there is more information in the position of the lips or in their movement ~\cite{luettin1996visual}, \cite{sahu2013result}, \cite{cappelletta2011viseme} and if visemes are better defined in terms of articulatory gestures (such as lips closing together, jaw movement, teeth exposure) or derived from the grouping of phonemes having the same visual appearance ~\cite{cappelletta2011viseme}, \cite{fisher1968confusions}. From a modeling viewpoint, the use of viseme units is essentially a form of model clustering that allows visually similar phonetic events to share a group model \cite{hilder2009comparison}. Consequently several different viseme vocabularies have been proposed in the literature typically with lengths between 11 and 15 visemes ~\cite{bear2014phoneme}, \cite{hazen2004segment}, \cite{potamianos2003recent}, \cite{neti2000audio}. For instance, Goldschen et al. \cite{goldschen1994continuous} trained an initial set of 56 phones and clustered them into 35 visemes using the Average Linkage hierarchical clustering algorithm. Jeffers et al. \cite{jeffers1980speechreading} defined a phoneme to viseme mapping from 50 phonemes to 11 visemes in the English language (11 visemes plus \textit{Silence}). Neti et al. \cite{neti2000audio} investigated the design of context questions based on decision trees to reveal similar linguistic context behaviour between phonemes that belong to the same viseme. For study, based on linguistic properties, they determined seven consonant visemes (bilabial, labio-dental, dental, palato-velar, palatal, velar, and two alveolar), four vowel, an alveolar-semivowel and one silence viseme (13 visemes in total). Bozkurt et al. \cite{bozkurt2007comparison} proposed a phoneme to viseme mapping from 46 American English phones to 16 visemes to achieve nature looking lip animation. They mapped phonetic sequences to viseme sequences before animating the lips of 3D head models. Ezzat et al. \cite{ezzat1998miketalk} presented a text-to-audiovisual speech synthesizer which converts input text into an audiovisual speech stream. They started grouping those phonemes which looked similar by visually comparing the viseme images. To obtain a photo-realistic talking face they proposed a viseme vocabulary with 6 visemes that represent 24 consonant phonemes, 7 visemes that represent the 12 vowel phonemes, 2 diphthong visemes and one viseme corresponding to the silence.

\subsection{Contributions}

In this work we investigate in the automatic construction of a viseme vocabulary from the association of visually similar phonemes. In contrast to the related literature, where visemes have been mainly defined manually (based on linguistic properties) or semi-automatically (e.g. by trees or clustering) \cite{cappelletta2011viseme} we explore the fully automatic construction of an optimal viseme vocabulary based on simple merging rules and the minimization of pair-wise confusion. We focus on constructing a VASR for Spanish language and explore the use of SIFT and DCT as descriptors for the mouth region, encoding both the spatial and temporal domains. We evaluated our system in a Spanish corpus (AV@CAR) with continuous speech from 20 speakers. Our results indicate that we are able to recognize more than 47\% of the phonemes and 23\% of the words corresponding to continuous speech and that the optimal viseme vocabulary for Spanish language is composed by 20 visemes.

\section{\uppercase{VASR system}}
\label{sec:Oursystem}
VASR systems typically aim at interpreting the video signal in terms of visemes, and usually consist of 3 major steps: 1) Lips localization, 2) Extraction of visual features, 3) Classification into viseme sequences. In this section we start with a brief review of the related work and then provide a detailed explanation of our method.

\subsection{Related Work}

Much of the research on VASR has focused on digit recognition, isolated words and sentences, and only more recently in continuous speech. Seymour et al. \cite{seymour2008comparison} centred their experiments in comparing different image transforms (DCT, DWT, FDCT) to achieve speaker-independent digit recognition. Sui et al. \cite{sui2015listening} presented a novel feature learning method using Deep Boltzmann Machines that recognizes simple sequences of isolated words and digit utterances. Their method used both acoustic and visual information to learn features, except for the test stage where only the visual information was used. Lan et al. \cite{lan2009comparing} used AAM features to quantify the effect of shape and appearance in lip reading and tried to recognize short sentences using a constrained vocabulary for 15 speakers. Zhao et al. \cite{zhao2009lipreading} proposed a spatiotemporal version of LBP features and used a SVM classifier to recognize isolated phrase sequences. Zhou et al. \cite{zhou2014compact} used a latent variable model that identifies two different sources of variation in images, those related to the appearance of the speaker and those caused by the pronunciation, and tried to separate them to recognize short utterances (e.g. \textit{Excuse me}, \textit{Thank you},...). Pet et al. \cite{pei2013unsupervised} presented a random forest manifold alignment method (RFMA) and applied it to lip-reading in color and depth videos. The lip-reading task was realized by motion pattern matching based on the manifold alignment. Potamianos et al. \cite{potamianos2003recent} applied fast DCT to the region of interest (ROI) and retained 100 coefficients. To reduce the dimensionality they used an intraframe linear discriminant analysis and maximum likelihood linear transform (LDA and MLLT), resulting in a 30-dimensional feature vector. To capture dynamic speech information, 15 consecutive feature vectors were concatenated, followed by an interframe LDA/MLLT for dimensionality reduction to obtain dynamic visual features of length 41. They tested their system using IBM ViaVoice database and reported 17.49\% of recognition rate in continuous speech recognition. Thangthai et al. \cite{thangthai2015improving} explored the use of Deep Neural Networks (DNNs) in combination with HiLDA features (LDA and MLLT). They reported very high accuracy ($\approx$ 85\%) in recognizing continuous speech although tests were on a corpus with a single speaker. Cappelletta et al. \cite{cappelletta2011viseme} used a database with short balanced utterances and tried to define a viseme vocabulary able to recognize continuous speech. They based their feature extraction on techniques as PCA or Optical flow, taking into account both movement and appearance of the lips.

Although some attempts to compare between methods have been made it is a quite difficult task in visual only ASR. Firstly, the recognition rates cannot be compared directly among recognition tasks: it is easier to recognize isolated digits trained with higher number of repetitions and number of speakers or to recognize shorts sentences trained in restricted vocabularies than to recognize continuous speech. Additionally, even when dealing with the same recognition tasks, the use of substantially different databases makes it difficult the comparison between methods. Concretely, results are often not comparable because they are usually reported in different databases, with variable number of speakers, vocabularies, language and so on. Keeping in mind these limitations, some studies have shown that most methods recognize automatically between 25\% and 64\% of short utterances \cite{zhou2014compact}. As mentioned before we are interested in continuous speech recognition because it is the task that is closer to actual lip-reading as done by humans. Continuous speech recognition has been explored recently, and there is limited literature about it. The complexity of the task and the few databases directly related to it have slowed its development, achieving rather low recognition rates. Because which technique use in each block of the pipeline is still an open problem, we decided to construct our own visual only ASR system based on intensity descriptors and on HMMs to model the dynamics of the speech.

\subsection{Our System}
\begin{figure*}[t]
\begin{center}
\includegraphics[width=\linewidth]{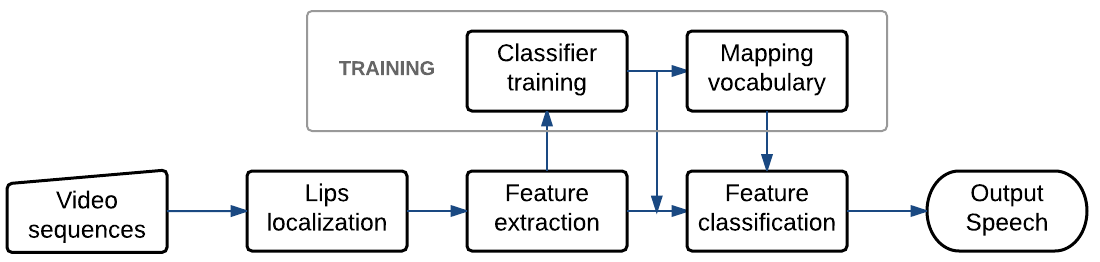}
\caption{General process of a VASR system.}
%; (b) IOF-ASM detection, the marks in blue are used to fix the bounding box; (c) ROI detection, each color fix a lateral of the bounding box.}
\label{fig:process}
\end{center}
\end{figure*}
In this section each step of our VASR system is explained (Figure~\ref{fig:process}). We start by detecting the face and extracting a region of interest (ROI) that comprises the mouth and its surrounding area. Appearance features are then extracted and used to estimate visemes, which are finally mapped into phonemes with the help of HMMs.

\subsubsection{Lips Localization}
\begin{figure}
\begin{center}
\begin{tabular}{cc}
{\fbox{\includegraphics[width=3cm, height=2.8cm]{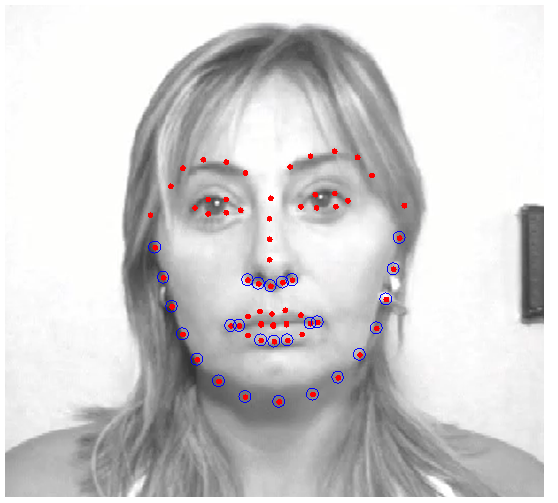}}} &
{\fbox{\includegraphics[width=3cm, height=2.8cm]{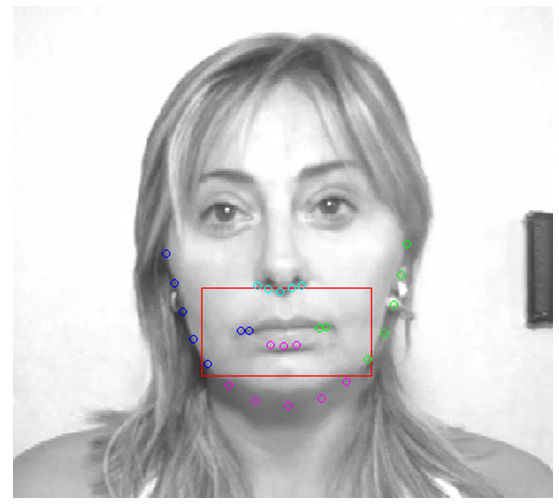}}} \\
(a)&(b) \\
\\
\end{tabular}
\caption{(a) IOF-ASM detection, the marks in blue are used to fix the bounding box; (b) ROI detection, each color fix a lateral of the bounding box.}
\label{fig:lipsLocalization}
\end{center}
\end{figure}

The location of the face is obtained using invariant optimal features ASM (IOF-ASM) \cite{sukno2007active} that provides an accurate segmentation of the face in frontal views. The face is tracked at every frame and detected landmarks are used to fix a bounding box around the lips (ROI) (Figure~\ref{fig:lipsLocalization}). At this stage the ROI can have a different size in each frame. Thus, ROIs are normalized to a fixed size of $48 \times 64$ pixels to achieve a uniform representation.

\begin{figure}
\begin{center}
\includegraphics[width=4cm, height=2.6cm]{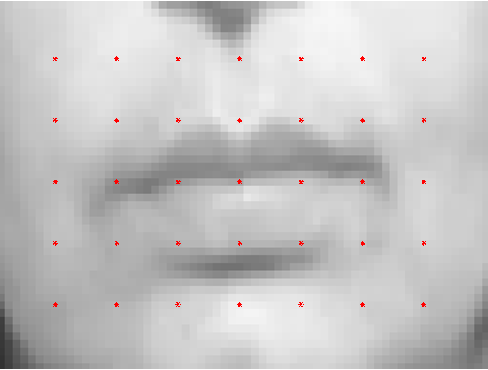}
\caption{Keypoints distribution.}
\label{fig:Keypointsdistribution}
\end{center}
\end{figure}

\subsubsection{Feature Extraction}
After the ROI is detected a feature extraction stage is performed. Nowadays, there is no universal feature for visual speech representation in contrast to the Mel-frequency cepstral coefficients (MFCC) for acoustic speech. We look for an informative feature invariant to common video issues, such as noise or illumination changes. We analyze three different appearance-based techniques:

\begin{itemize}
\item \textit{SIFT}: SIFT was selected as high level descriptor to extract the features in both the spatial and temporal domains because it is highly distinctive and invariant to image scaling and rotation, and partially invariant to illumination changes and 3D camera viewpoint \cite{lowe2004distinctive}. In the spatial domain, the SIFT descriptor was applied directly to the ROI, while in the temporal domain it was applied to the centred gradient. SIFT keypoints are distributed uniformly around the ROI (Figure~\ref{fig:Keypointsdistribution}). The distance between keypoints was fixed to half of the neighbourhood covered by the descriptor to gain robustness (by overlapping). As the dimension of the final descriptor for both spatial and temporal domains is very high, PCA was applied to reduce the dimensionality of the features. Only statistically significant components (determined by means of Parallel Analysis \cite{franklin1995parallel}) were retained.
\item \textit{DCT}: The 2D DCT is one of the most popular techniques for feature extraction in visual speech \cite{zhou2014review}, \cite{lan2009comparing}. Its ability to compress the relevant information in a few coefficients results in a descriptor with small dimensionality. The 2D DCT was applied directly to the ROI. To fix the number of coefficients, the image error between the original ROI and the reconstructed was used. Based on preliminary experiments, we found that 121 coefficients (corresponding to 1\% reconstruction error) for both the spatial and temporal domains produced satisfactory performance.
\item \textit{PCA}: Another popular technique is PCA, also known as eigenlips ~\cite{zhou2014review}, \cite{lan2009comparing}, \cite{cappelletta2011viseme}. PCA, as 2D DCT is applied directly to the ROI. To decide the optimal number of dimensions the system was trained and tested taking different percentages of the total variance. Lower number of components would lead to a low quality reconstruction, but an excessive number of components will be more affected by noise. In the end 90\% of the variance was found to be a good compromise and was used in both spatial and temporal descriptors.
\end{itemize}

The early fusion of DCT-SIFT and PCA-SIFT has been also explored to obtain a more robust descriptor (see results in Section~\ref{sec:results}).

\subsubsection{Feature Classification and Interpretation}
The final goal of this block is to convert the extracted features into phonemes or, if that is not possible, at least into visemes. To this end we need: 1) classifiers that will map features to (a first estimate of) visemes; 2) a mapping between phonemes and visemes; 3) a model that imposes temporal coherency to the estimated sequences.

\begin{enumerate}
\item{\textbf{Classifiers:}} classification of visemes is a challenging task, as it has to deal with issues such as class imbalance and label noise. Several methods have been proposed to deal with these problems, the most common solutions being Bagging and Boosting algorithms \cite{khoshgoftaar2011comparing}, \cite{verbaeten2003ensemble}, \cite{frenay2014classification}, \cite{nettleton2010study}. From these, Bagging has been reported to perform better in the presence of training noise and thus it was selected for our experiments. Multiple LDA was evaluated using cross validation. To add robustness to the system, we trained classifiers to produce not just a class label but to estimate also a class probability for each input sample.

For each bagging split, we train a multi-class LDA classifier and use the Mahalanobis distance $d$ to obtain a normalized projection of the data into each class $c$:
\begin{equation}
    d_c(x) = \sqrt{(x-\bar{x_c})^T \cdot \Sigma_c^{-1} \cdot (x-\bar{x_c})}
    \label{Eq:Mahalanobis}
\end{equation}
Then, for each class, we compute two cumulative distributions based on these projections: one for in-class samples $\Phi(\frac{d_c(x)-\mu_c}{\sigma_c}),\,x \in c$ and another one for out-of-class samples $\Phi(\frac{d_c(x)-\mu_{\widetilde{c}}}{\sigma_{\widetilde{c}}}),\,x \in \widetilde{c}$, which we assume Gaussian with means $\mu_c$, $\mu_{\widetilde{c}}$ and variances $\sigma_c$, $\sigma_{\widetilde{c}}$, respectively. An indicative example is provided in Figure~\ref{fig:gaussian&cdf}. Notice that these means and variances correspond to the projections in (\ref{Eq:Mahalanobis}) and are different from $\bar{x_c}$ and $\Sigma_c$.

We compute a class-likelihood as the ratio between the in-class and the out-of-class distributions, as in (\ref{Eq:likelihood}) and normalize the results so that the summation over all classes is 1, as in (\ref{Eq:likelihoodNorm}). When classifying a new sample, we use the cumulative distributions to estimate the probability that the unknown sample belongs to each of the viseme classes (\ref{Eq:likelihoodNorm}). We assign the class with the highest normalized likelihood $L_c$.

\begin{equation}
    F(c \mid x ) = \frac{1-\Phi(\frac{d_c(x)-\mu_c}{\sigma_c})}{ \Phi(\frac{d_c(x)-\mu_{\widetilde{c}}}{\sigma_{\widetilde{c}}})}
    \label{Eq:likelihood}
\end{equation}
\begin{equation}
    L_c(x) = \frac{F(c \mid x)}{\sum_{c=1}^{C} F(c \mid x)}
    \label{Eq:likelihoodNorm}
\end{equation}

\begin{figure}[tb]
\centering
\begin{minipage}[b]{1\linewidth}
\centering
\includegraphics[width = 7cm, height=4.5cm]{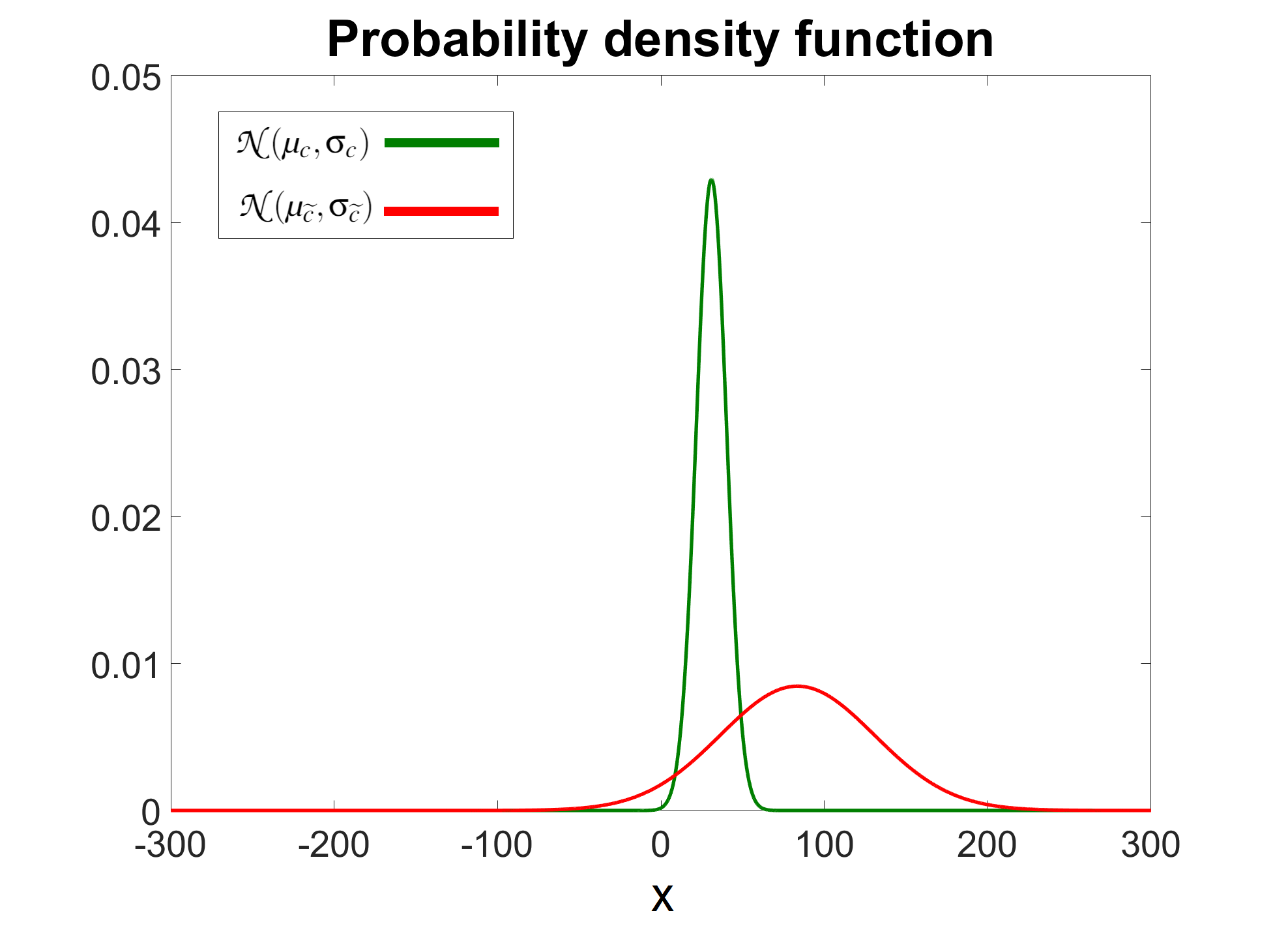}
\end{minipage}

\begin{minipage}[b]{1\linewidth}
\centering
\includegraphics[width = 7cm,height=4.5cm]{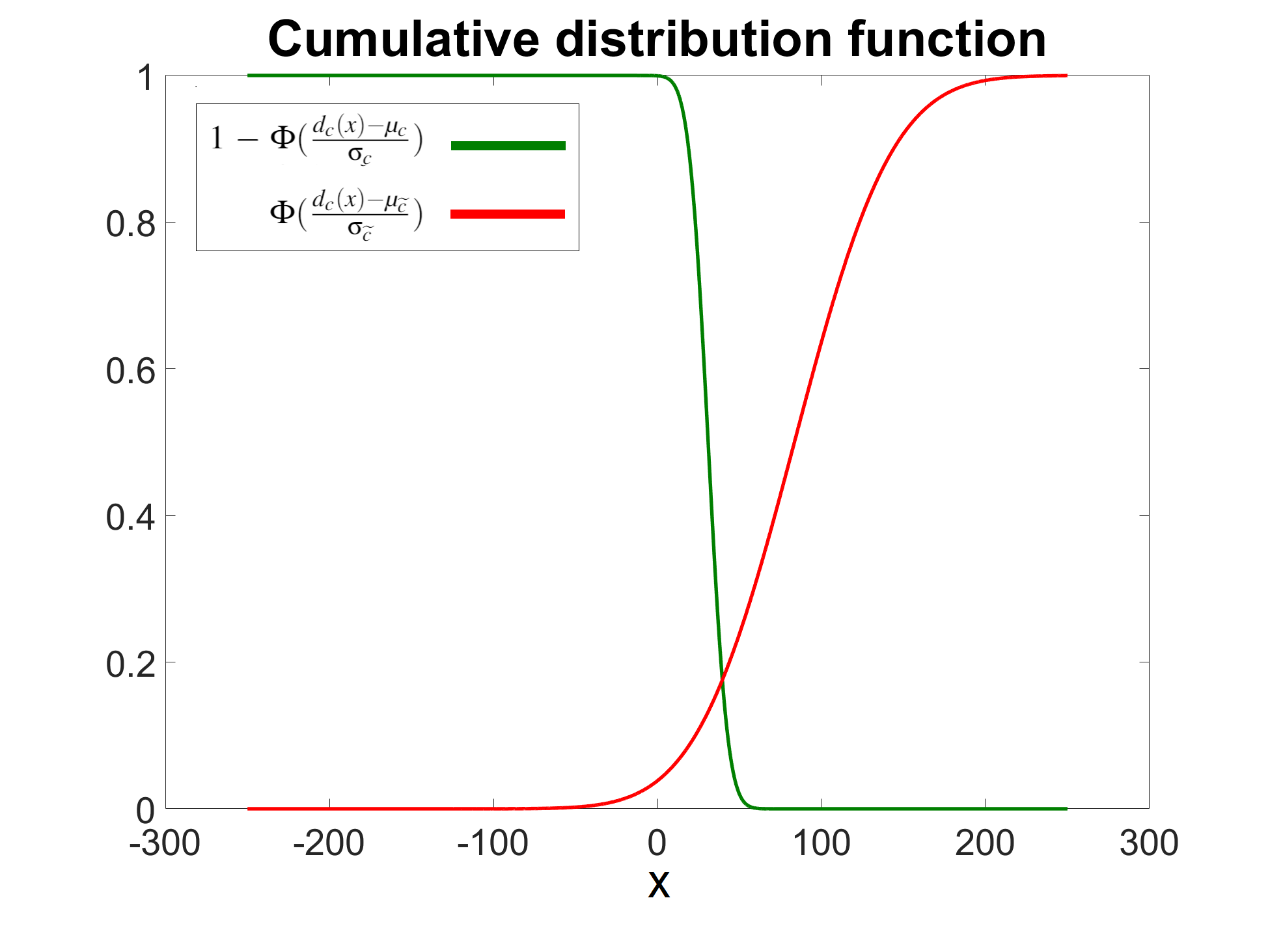}
\end{minipage}

\caption{ (\textit{Top}) Probability density functions for in-class (green) and out-of-class (red) samples; (\textit{Bottom}) Cumulative distributions corresponding to (\textit{Top}). Notice than for in-class samples we use the complement of the cumulative distribution, since lower values should have higher probabilities.}
\label{fig:gaussian&cdf}
\end{figure}

Once the classifiers are trained we could theoretically try to classify features directly into phonemes, but as explained in Section \ref{sec:intro}, there are phonemes that share the same visual appearance and are therefore unlikely to be distinguishable by a visual-only system. Thus, such phonemes should be grouped into the same class (visemes). In the next subsection we will present a mapping from phonemes to visemes based on grouping phonemes that are visually similar.

\item{\textbf{Phoneme to Viseme Mapping:}} to construct our viseme to phoneme mapping we analyse the confusion matrix resulting by comparing the ground truth labels of the training set with the automatic classification obtained from the previous section. We use an iterative process, starting with the same number of visemes as phonemes, merging at each step the visemes that show the highest ambiguity. The method takes into account that vowels cannot be grouped with consonants, because it has been demonstrated that their aggregation produces worse results \cite{cappelletta2011viseme}, \cite{bear2014phoneme}.

The algorithm iterates until the desired vocabulary length is achieved. However, there is no accepted standard to fix this value beforehand. Indeed, several different viseme vocabularies have been proposed in the literature typically with lengths between 11 and 15 visemes. Hence, in Section \ref{sec:results} we will analyse the effect of the vocabulary size on recognition accuracy. Once the vocabulary construction is concluded, all classifiers are retrained based on the resulting viseme classes.

\item{\textbf{HMM and Viterbi Algorithm:}} to improve the performance obtained after feature classification, HMMs of one state per class are used to map: 1) visemes to visemes; 2) visemes to phonemes.

An HMM $\lambda = (A, B, \pi)$ is formed by N states and M observations. Matrix A represents the state transition probabilities, matrix B the emission probabilities, and vector $\pi$ the initial state probabilities.
Given a sequence of observation O and the model $\lambda$ our aim is to find the maximum probability state path $Q = q_1, q_2, ..., q_{t-1}$. This can be done recursively using Viterbi algorithm \cite{rabiner1989tutorial}, \cite{petrushin2000hidden}. Let $\delta_i(t)$ be the probability of the most probable state path ending in state $i$ at time $t$ (\ref{Eq:viterbi}). Then $\delta_j(t)$ can be computed recursively using (\ref{Eq:viterbi2}) with initialization (\ref{Eq:viterbiInit}) and termination (\ref{Eq:vitermiTerm}).
\begin{equation}
    \delta_i(t) = \smash{\displaystyle\max_{q_1, \ldots, q_{t-1}}
    P (q_1 ... q_{t-1} = i, O_1,..., O_t | \lambda)}
    \label{Eq:viterbi}
\end{equation}

\begin{equation}
    \delta_j(t) = \smash{\displaystyle\max_{1 \leq i \leq N}} [\delta_i(t-1) \cdot a_{i,j}] \cdot b_j(O_t)
    \label{Eq:viterbi2}
\end{equation}

\begin{equation}
    \delta_i(1) = \pi_i \cdot b_i(O_1),    1 \leq i \leq N
    \label{Eq:viterbiInit}
\end{equation}

\begin{equation}
    P = \smash{\displaystyle\max_{1 \leq i \leq N}} [\delta_i(T)]
    \label{Eq:vitermiTerm}
\end{equation}

A shortage of the above is that it only considers a single observation for each instant $t$. In our case observations are the output from classifiers and contain uncertainty. We have found that it is useful to consider multiple possible observations for each time step. We do this by adding to the Viterbi algorithm the likelihoods obtained by the classifiers for all classes (e.g. from equation (\ref{Eq:likelihoodNorm})). As a result, (\ref{Eq:viterbi2}) is modified into (\ref{Eq:viterbiFinal}), where the maximization is done across both the $N$ states (as in (\ref{Eq:viterbi2})) and also the $M$ possible observations, each weighted with its likelihood estimated by the classifiers.
\begin{equation}
    \delta_{j}(t) = \smash{\displaystyle\max_{1 \leq O_t \leq M}}\;\smash{\displaystyle\max_{1 \leq i \leq N}} [\delta_i(t-1) \cdot a_{i,j}] \cdot \hat{b}_j(O_t)
    \label{Eq:viterbiFinal}
\end{equation}

\begin{equation}
    \hat{b}_j(O_t) = b_j(O_t) \cdot L(O_t)
\end{equation}
where the short-form $L(O_t)$ refers to the likelihood $L_{O_t}(x)$ as defined in (\ref{Eq:likelihoodNorm}). The Viterbi algorithm modified as indicated in (\ref{Eq:viterbiFinal}) is used to obtain the final viseme sequence providing at the same time temporal consistency and tolerance to classification uncertainties. Once this has been achieved, visemes are mapped into phonemes using the traditional Viterbi algorithm (\ref{Eq:viterbi2}).
\end{enumerate}

\section{\uppercase{Experiments}}
\subsection{Database}

\begin{table}
\begin{center}
\caption{Sentences for speaker 1.}
\label{tab:frases}
\begin{tabular}{|p{7cm}|}
 \hline
 \textbf{Speaker 1} \\
 \hline\hline
Francia, Suiza y Hungr\'{i}a ya hicieron causa com\'{u}n. \\
Despu\'{e}s ya se hizo muy amiga nuestra. \\
Los yernos de Ismael no engordar\'{a}n los pollos con hierba. \\
Despu\'{e}s de la mili ya me vine a Catalu\~{n}a. \\
Bajamos un d\'{i}a al mercadillo de Palma. \\
Existe un viento del norte que es un viento fr\'{i}o. \\
Me he tomado un caf\'{e} con leche en un bar. \\
Yo he visto a gente expulsarla del colegio por fumar. \\
Guadalajara no est\'{a} colgada de las rocas. \\
Pas\'{e} un a\~{n}o dando clase aqu\'{i}, en Bellaterra. \\
Pero t\'{u} ahora elijes ya previamente. \\
Les dijeron que eligieran una casa all\'{a}, en las mismas condiciones. \\
Cuando me gir\'{e} ya no ten\'{i}a la cartera. \\
Tendr\'{a} unas siete u ocho islas alrededor. \\
Haciendo el primer campamento y el segundo campamento. \\
Unas indemnizaciones no les iban del todo mal. \\
Rezando porque ten\'{i}a un miedo impresionante. \\
Es un apellido muy abundante en la zona de Pamplona. \\
No jug\'{a}bamos a b\'{a}sket, s\'{o}lo los mir\'{a}bamos a ellos. \\
Aunque naturalmente hay un partido comunista. \\
Dio la casualidad que a la una y media estaban all\'{i}. \\
Se alegraron mucho de vernos y ya nos quedamos a cenar. \\
Ya empezamos a llorar bastante en el apartamento. \\
En una ladera del monte se ubica la iglesia. \\
Entonces lo \'{u}nico que hac\'{i}amos era ir a cenar. \\
 \hline
\end{tabular}
\end{center}
\end{table}

Ortega et al. \cite{ortega2004av} introduced AV@CAR as a free multichannel multi-modal database for automatic audio-visual speech recognition in Spanish language, including both studio and in-car recordings. The Audio-Visual-Lab data set of AV@CAR contains sequences of 20 people recorded under controlled conditions while repeating predefined phrases or sentences. There are 197 sequences for each person, recorded in AVI format. The video data has a spatial resolution of 768x576 pixels, 24-bit pixel depth and 25 fps and is compressed at an approximate rate of 50:1. The sequences are divided into 9 sessions and were captured in a frontal view under different illumination conditions and speech tasks. Session 2 is composed by 25 videos/user with phonetically-balanced phrases. We have used session 2 splitting the dataset in 380 sentences (19 users $\times$ 20 sentences/user) for training and 95 sentences (19 users $\times$ 5 sentences/user) to test the system. In Table \ref{tab:frases} it is shown the sentences of the first speaker.

\subsection{Phonetic Vocabulary}
SAMPA is a phonetic alphabet developed in 1989 by an international group of phoneticians, and was applied to European languages as Dutch, English, French, Italian, Spanish, etc. We based our phonetic vocabulary in SAMPA because it is the most used standard in phonetic transcription \cite{wells1997sampa}, \cite{llisterri1993spanish}. For the Spanish language, the vocabulary is composed by the following 29 phonemes: /p/, /b/, /t/, /d/, /k/, /g/, /tS/, /jj/, /f/, /B/, /T/, /D/, /s/, /x/, /G/, /m/, /n/, /J/, /l/, /L/, /r/, /rr/, /j/, /w/, /a/, /e/, /i/, /o/, /u/. The phonemes /jj/ and /G/ were removed from our experiments because our database did not contain enough samples to consider them.

\subsection{Results}
\label{sec:results}
In this section we show the results of our experiments. In particular, we show the comparison of the performances between the different vocabularies, the different features, and the improvement obtained by adding the observation probabilities into the Viterbi algorithm.

\subsubsection{Experimental Setup}
We constructed an automatic system that uses local appearance features based on early fusion of DCT and SIFT descriptors (this combination produced the best results in our tests, see below) to extract the main characteristics of the mouth region in both spatial and temporal domains. The classification of the extracted features into phonemes is done in two steps. Firstly, 100 LDA classifiers are trained using bagging sequences to be robust under label noise. Then, the classifier outputs are used to compute the global normalized likelihood, as the summation over the normalized likelihood computed by each classifier divided by the number of classifiers (as explained in Section \ref{sec:Oursystem}). Secondly, at the final step, one-state-per-class HMMs are used to model the dynamic relations of the estimated visemes and produce the final phoneme sequences.

\subsubsection{Comparison of Different Vocabularies}
\begin{figure}[tb]
\begin{center}
\includegraphics[width=7cm,height=5cm]{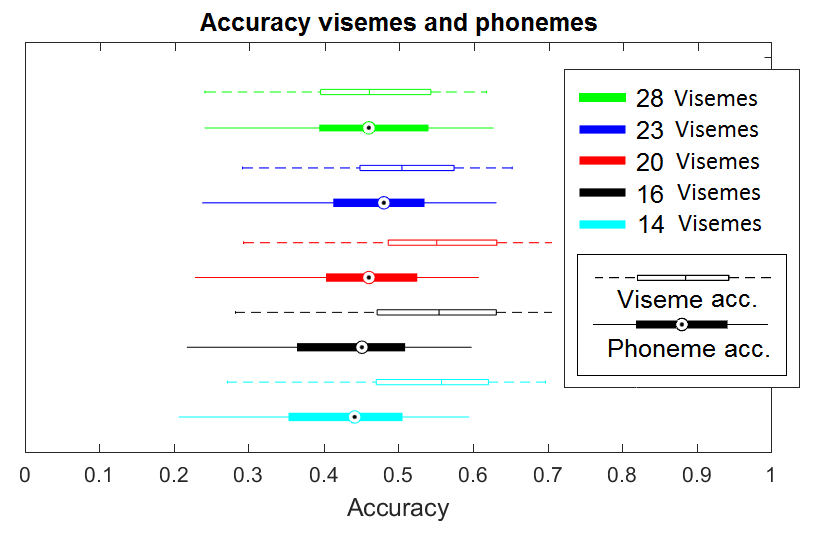}
\caption{Boxplots of system performance in terms of phoneme and viseme accuracy for different vocabularies. We analyze the one-to-one mapping phoneme to viseme, and the one-to-many phoneme to viseme mappings with 23, 20, 16 and 14 visemes. The phoneme accuracy is always computed from the 28 phonemes.}
\label{fig:VisemeLengthComparison}
\end{center}
\end{figure}

\begin{figure}[tb]
\begin{center}
\includegraphics[width=\linewidth, height=3.8cm]{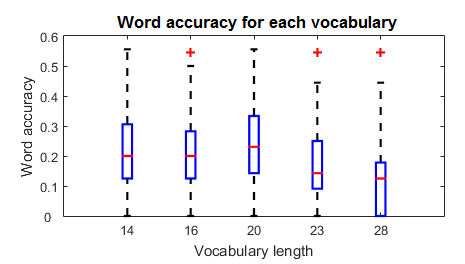}
\caption{Comparison of system performance in word recognition rate for the different vocabularies.}
\label{fig:VisemeLengthComparisonWacc}
\end{center}
\end{figure}

As we explained before one of the main challenges of VASR systems is the definition of the phoneme-to-viseme mapping. While our system aims to estimate phoneme sequences, we know that there is no one-to-one mapping between phonemes and visemes. Hence, we try to find the one-to-many mapping that will allow us to maximize the recognition of phonemes in the spoken message.

To evaluate the influence of the different mappings, we have analysed the performance of the system in terms of viseme- , phoneme-, and word recognition rates using viseme vocabularies of different lengths.
Our first observation, from Figure \ref{fig:VisemeLengthComparison}, is that the viseme accuracy tends to grow as we reduce the vocabulary length. This is explained by two factors: 1) the reduction in number of classes, which makes the classification problem a simpler one to solve; 2) the fact that visually indistinguishable units are combined into one. The latter helps explain the behaviour of the other metric in the figure: phoneme accuracy. As we reduce the vocabulary length, phoneme accuracy firstly increases because we eliminate some of the ambiguities by merging visually similar units. But if we continue to reduce the vocabulary, too many phonemes (even unrelated) are mixed together and their accuracy decreases because, even if these visemas are better recognized, their mapping into phonemes is more uncertain.
Thus, the optimal performance is obtained for intermediate vocabulary lengths, because there is an optimum compromise between the visemes and the phonemes that can be recognized.

The same effect can also be seen in Figure \ref{fig:VisemeLengthComparisonWacc} in terms of word recognition rates. We can observe how the one-to-one phoneme to viseme mapping (using the 28 phonemes classes) obtained the lowest word recognition rates and how the highest word recognition rates were obtained for the intermediate vocabulary lengths, supporting the view that the one-to-many mapping from phonemes to visemes is necessary to optimize the performance of visual speech systems. In the experiments presented in this paper, a vocabulary of 20 visemes (summarized in Table \ref{tab:visemeVoc}) produced the best performance.

\begin{table}
\begin{center}
\caption{Optimal vocabulary obtained, composed of 20 visemes.}
\label{tab:visemeVoc}
\begin{tabular}{|c|c|c|c|}
 \hline
 \multicolumn{4}{|c|}{\textbf{Viseme vocabulary}} \\
 \hline\hline
 Silence & T &  l & rr\\
 \hline
 D & ll & d & s, tS, t\\
 \hline
 k & x & j & r \\
 \hline
 g  &  m, p, b & f & n  \\
 \hline
 B & J & a,e,i  & o,u,w  \\
 \hline
\end{tabular}
\end{center}
\end{table}

\subsubsection{Feature Comparison}
\begin{figure}[tb]
\centering
\begin{minipage}[b]{1\linewidth}
\centering
\includegraphics[height=4cm]{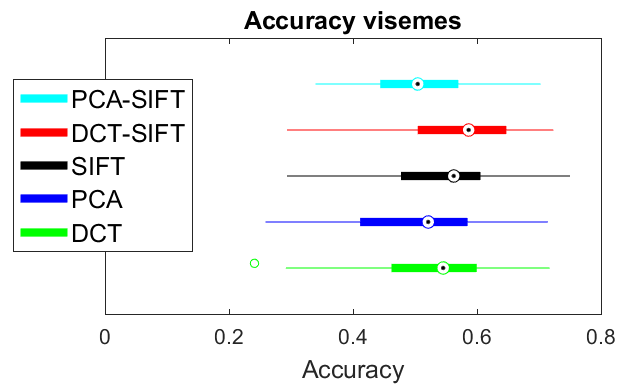}
\end{minipage}
\quad
\begin{minipage}[b]{1\linewidth}
\centering
\includegraphics[height=4cm]{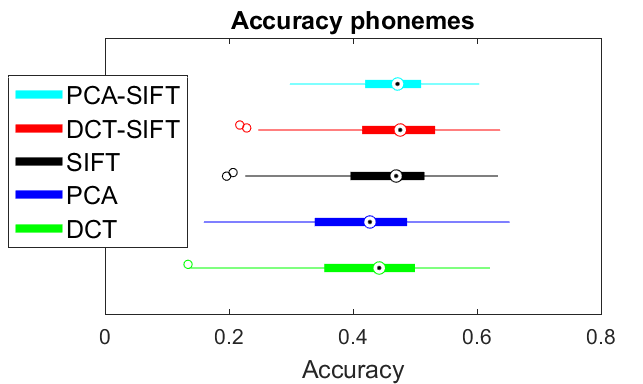}
\end{minipage}
\caption{Comparison of features performance.}
\label{fig:FeaturesComparison}
\end{figure}

To analyse the performance of the different features we have fixed the viseme vocabulary as shown in Table~\ref{tab:visemeVoc} and performed a 4-fold cross-validation on the training set. We used 100 LDA classifiers per fold, generated by means of a bagging strategy. Figure~\ref{fig:FeaturesComparison} displays the results. Visualizing the features independently, DCT and SIFT give the best performances. The fusion of both features produced an accuracy of 0.58 for visemes, 0.47 for phonemes.

\subsubsection{Improvement by Adding Classification Likelihoods}

\begin{figure}[tb]
\centering
\includegraphics[width=\linewidth, height=3.5cm]{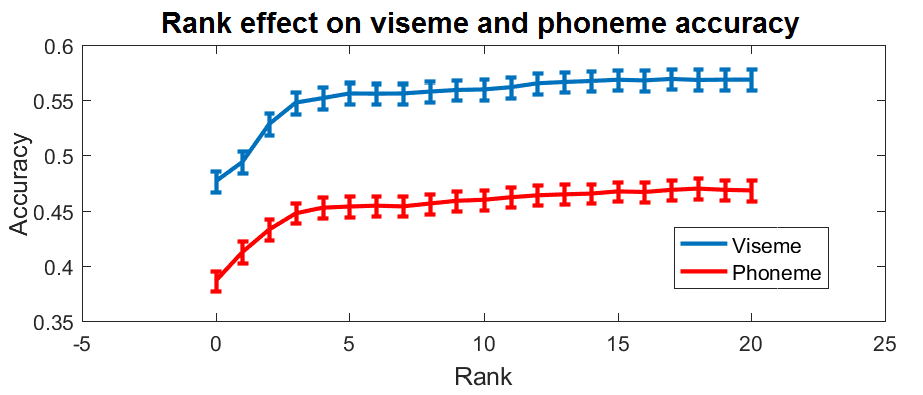}
\caption{Accuracy for different ranks in the Viterbi algorithm.}
\label{fig:rankViterbi}
\end{figure}

Figure~\ref{fig:rankViterbi} shows how the accuracy varies when considering the classifier likelihoods in the Viterbi algorithm. The horizontal axis indicates the number of classes that are considered (the rank), in decreasing order of likelihood. The performance of the algorithm without likelihoods (\ref{Eq:viterbi2}) is also provided as a baseline (rank 0). We see that the improvement obtained by the inclusion of class likelihoods is up to 20\%.

\begin{figure*}[ht]
\centering
\includegraphics[width=13.3cm, height = 11.5cm]{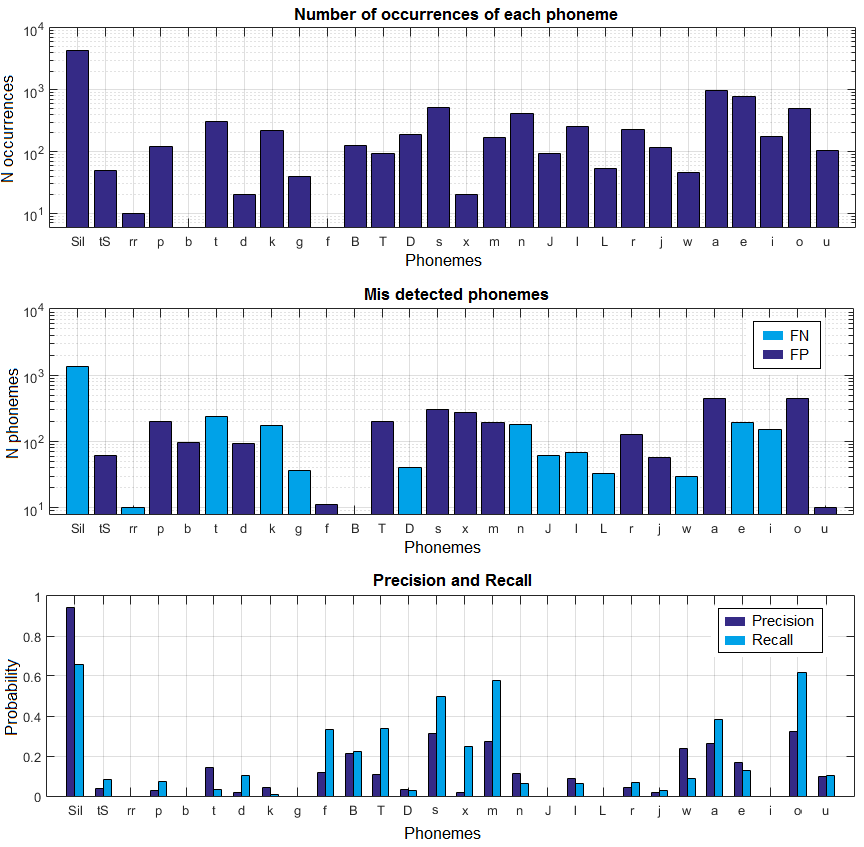}
\caption{(a) Number of occurrences of each viseme; (b) Visemes wrongly detected (actual - detected), light blue (false negatives) and dark blue (false positives); (c) Precision and Recall of visemes.}
\label{fig:PRMisdetected}
\end{figure*}

Finally, it is interesting to analyse how the system performs for each of the resulting phonemes. Figure~\ref{fig:PRMisdetected} (a) shows the frequency of appearance of each phoneme. In Figure~\ref{fig:PRMisdetected} (b) we show the number of phonemes that are wrongly detected. It can be seen that the input data is highly unbalanced, biasing the system toward the phonemes that appear more often. For example, the \textit{silence} appears 4 times more than the vowels \textit{a,e,i} and there are some phonemes with very few samples, such as \textit{rr}, \textit{f} or \textit{b}. This has also an impact in terms of precision and recall, as can be observed in Figure~\ref{fig:PRMisdetected} (c). In the precision and recall figure we can observe the effects of the many-to-one viseme to phoneme mapping. For example, there are phonemes with low precision and recall because have been confused with one of the phonemes of their viseme group (e.g: vowel \textit{i} have been confused with \textit{a and e}). Considering the three plots at once, there is a big impact of the silence on the overall performance of the system. In particular, the recognition of silences shows a very high precision but its recall is only about 70\%. By inspecting our data, we found that this is easily explained by the fact that, normally, people start moving their lips before speaking, in preparation for the upcoming utterance. Combination with audio would easily resolve this issue.

\section{\uppercase{Conclusions}}
We investigate the automatic construction of optimal viseme vocabularies by iteratively combining phonemes with similar visual appearance into visemes. We perform tests on the Spanish database AV@CAR using a VASR system based on the combination of DCT and SIFT descriptors in spatial and temporal domains and HMMs to model both viseme and phoneme dynamics. Using 19 different speakers we reach a 58\% of recognition accuracy in terms of viseme units, 47\% in terms of phoneme units and 23\% in terms of words units, which is remarkable for a multi-speaker dataset of continuous speech as the one used in our experiments.
Comparing the performance obtained by our viseme vocabulary with the performance of other vocabularies, such as those analysed by Cappelleta et al. in \cite{cappelletta2011viseme}, we observe that the 4 vocabularies they propose have lengths of 11, 12, 14 and 15 visemes and their maximum accuracy is between 41\% and 60\% (in terms of viseme recognition).

Interestingly, while our results support the advantage of combining multiple phonemes into visemes to improve performance, the number of visemes that we obtain are comparatively high with respect to previous efforts. In our case, the optimal vocabulary length for Spanish reduced from $28$ phonemes to $20$ visemes (including $Silence$), i.e. a reduction rate of about $3:2$. In contrast, previous efforts reported for English started from $40$ to $50$ phonemes and merged them into just $11$ to $15$ visemes \cite{cappelletta2011viseme}, with reduction rates from $3:1$ to $5:1$. It is not clear, however, if the higher compression of the vocabularies obeys to a difference inherent to language or to other technical aspects, such as the ways of defining the phoneme to viseme mapping.

Indeed, language differences make it difficult to make a fair comparison of our results with respect to previous work. Firstly, it could be argued that our viseme accuracy is comparable to values reported by Cappelletta et al. \cite{cappelletta2011viseme}; however they used at most $15$ visemes while we use $20$ visemes and, as shown in Figure~\ref{fig:VisemeLengthComparison}, when the number of visemes decreases, viseme recognition accuracy increases but phoneme accuracy might be reduced hence making more difficult to recover the spoken message. Unfortunately, Cappelletta et al. \cite{cappelletta2011viseme} did not report their phoneme recognition rates.

Another option for comparison is word-recognition rates, which are frequently reported in automatic speech recognition systems. However, in many cases recognition rates are reported only for audio-visual systems without indicating visual-only performance  \cite{hazen2004segment}, \cite{cooke2006audio}. Within systems reporting visual-only performance, comparison is also difficult given that they are often centered on tasks such as digit or sentence recognition \cite{seymour2008comparison}, \cite{sui2015listening}, \cite{zhao2009lipreading}, \cite{saenko2005visual}, which are considerably simpler than the recognition of continuous speech, as addressed here. Focusing on continuous systems for visual speech, Cappelleta et al. \cite{cappelletta2011viseme} did not report word recognition rates and Thangthai et al. \cite{thangthai2015improving} reported tests just on a single user. Finally, Potamianos et al. \cite{potamianos2003recent} implemented a system comparable to ours based on appearance features and tested it using the multi-speaker IBM ViaVoice database, achieving 17.49\% of word recognition rate in continuous speech, which is not far from the 23\% of word recognition rate achieved by our system.

\section*{ACKNOWLEDGEMENTS}
This work is partly supported by the Spanish Ministry of Economy and Competitiveness under the Ramon y Cajal fellowships and the Maria de Maeztu Units of Excellence Programme (MDM-2015-0502), and the Kristina project funded by the European Union Horizon 2020 research and innovation programme under grant agreement No 645012.
%\clearpage
\vfill
{\small
%\bibliography{example}

}

\vfill
\end{document}